\documentclass[letterpaper, 10 pt, conference]{ieeeconf}  %

\IEEEoverridecommandlockouts                              %

\usepackage{url}
\usepackage{graphicx} %
\usepackage{booktabs}
\usepackage{helvet}   %
\usepackage{tikz}
\usetikzlibrary{positioning,arrows.meta,fit,backgrounds,calc,decorations.pathreplacing}
\usepackage{amsmath} %
\usepackage{amssymb}  %
\usepackage{balance}
\usepackage{makecell}

\makeatletter
\let\NAT@parse\undefined
\makeatother

\usepackage{cite}
\usepackage[colorlinks,allcolors=magenta]{hyperref} %
\usepackage[all]{hypcap}

\title{\LARGE \bf
HexVIO: Towards All-Day Stereo-Inertial Tracking Through Commodity DSPs
}

\author{Patrick Wolf$^{*1}$, Mateo de Mayo$^{*1,2}$, Daniel Cremers$^{1,2}$%
\thanks{*Equal contribution.}%
\thanks{$^{1}$Patrick Wolf, Mateo de Mayo, and Daniel Cremers are with the Technical University of Munich, Munich, Germany.
        {\tt\small \{pm.wolf, mateo.demayo, cremers\}@tum.de}}%
\thanks{$^{2}$Mateo de Mayo and Daniel Cremers are with the Munich Center for Machine Learning, Munich, Germany.}%
}

\newread\imgstream
\immediate\openin\imgstream=imagedata.in
\makeatletter
\def\new@kvginclip#1{}
\def\new@kvgintrim#1{}
\let\old@kvginclip\KV@Gin@clip
\let\old@kvgintrim\KV@Gin@trim
\let\oldincludegraphics\includegraphics
\providecommand{\includegraphics}{}
\renewcommand{\includegraphics}[2][]{%
  \immediate\read\imgstream to \src
  \immediate\read\imgstream to \removecrop
  \ifnum\removecrop=1
      \let\KV@Gin@clip\new@kvginclip
      \let\KV@Gin@trim\new@kvgintrim
  \fi
  \oldincludegraphics[#1]{\src}%
  \let\KV@Gin@clip\old@kvginclip
  \let\KV@Gin@trim\old@kvgintrim}
\makeatother

\begin{document}

\maketitle
\thispagestyle{empty}
\pagestyle{empty}

\begin{abstract}

The ability of a device to localize itself within its surroundings is a fundamental prerequisite for spatial computing. Visual-inertial odometry (VIO) has proven to be a cost-effective and accurate solution for this task. 
Robots, wearables, XR devices, and drones can benefit significantly from efficient implementations of VIO since they allow for cooler, lighter, and cheaper devices with longer battery life and a better user experience. In this work, we propose to enhance the efficiency of a VIO system by leveraging the Hexagon DSP, a commodity co-processor present in many modern smartphones and XR devices. Our approach offloads the visual frontend of a stereo-inertial odometry system to the DSP while keeping the backend on the main CPU. By optimizing the implementation for the DSP architecture, we achieve significant reductions in power consumption and latency compared to CPU-only execution. Our system, HexVIO, demonstrates a 67\% reduction in power consumption or an 86\% increase in throughput on a commodity smartphone, with the ability to sustain long-term real-time 30\,fps tracking for 0.83\,W, corresponding to $\sim$18 hours of tracking on the testing device. These results highlight the potential of commodity DSPs for enabling all-day visual-inertial tracking in robotics and mobile devices.

\end{abstract}

\section{INTRODUCTION}

\begin{figure*}[t]
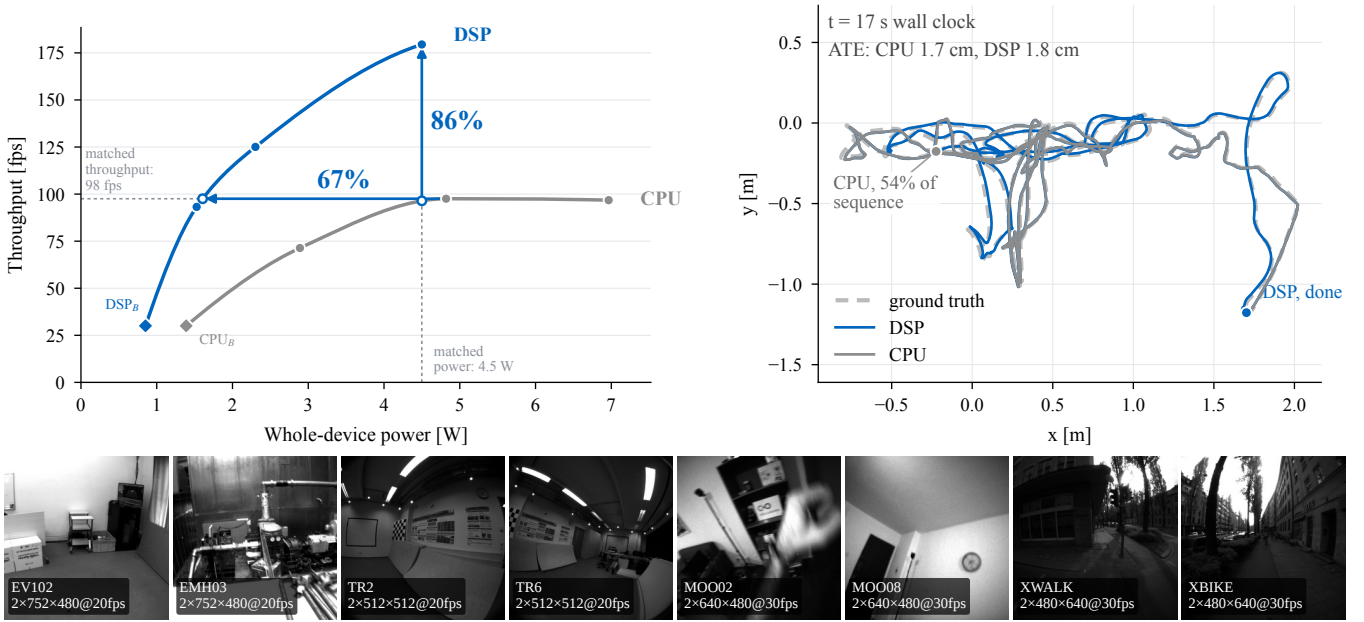

    \centering
    \includegraphics{figures/teaser_scaling.pdf}\hfill
    \includegraphics{figures/teaser_traj.pdf}\\[2pt]
    \includegraphics[width=\textwidth]{figures/teaser_images.pdf}
    \caption{Top left: efficiency of HexVIO (DSP) against the CPU baseline on a Samsung Galaxy S23, MOO05 sequence. Each curve spans four operating points of one implementation: the paced 30\,fps configuration (CPU$_B$, DSP$_B$), the frontend on two and on four mid-cluster threads, and a fourth point that is unpinned for the CPU and the maximum-performance RPC policy of Table~\ref{tab:performance} for the DSP. At the baseline's maximum throughput (98\,fps), HexVIO draws 67\,\% less power; at HexVIO's maximum-performance point (4.5\,W), it delivers 86\,\% higher throughput than the CPU at the same power. Filled markers are measured operating points (mean of two runs); curves are shape-preserving cubic interpolants (PCHIP), and open markers are matched points read from the interpolants. Top right: top-down trajectories of both implementations on the same sequence against ground truth, frozen when the unpaced DSP run finishes; the unpaced CPU run has processed 54\,\% of the sequence at the same wall-clock time, with the same ATE (1.7 vs 1.8\,cm). Bottom: one frame from each of the eight evaluation sequences, labeled with camera count, resolution and frame rate.}
    \label{fig:teaser}
\end{figure*}

\setcounter{footnote}{0}

Recent advancements in robotics \cite{zitkovichRT2VisionLanguageActionModels2023,zhangHuBLearningExtreme2025}, computer vision \cite{wangVGGTVisualGeometry2025a,ICLR2026_e0982cbc,ICLR2025_bc8b2058,kerblImpactOutlook3D2025}, and foundation models capable of complex reasoning \cite{openaiGPT4TechnicalReport2024,guoDeepSeekR1IncentivizesReasoning2025,chowdheryPaLMScalingLanguage2023} have started to show the possibilities of perception engineering \cite{lavalleVirtualRealityEmerging2024a} as a new discipline and of spatial computing \cite{delmericoSpatialComputingIntuitive2022} as the ever more common class of problems that require a computing device to understand its surroundings. Different instances of this problem arise in important applications like autonomous vehicles, drones, humanoid robotics, wearables like smartglasses, and VR headsets. A fundamental capability that allows much of this reasoning comes from the ability of the device to localize itself and its movements in space. Utilizing cameras and inertial measurement units (IMU) for \textit{visual-inertial odometry} (VIO) and \textit{visual-inertial simultaneous localization and mapping} (VI-SLAM) has shown to be a particularly good combination thanks to the low cost of these sensors, their reduced power consumption, and proven accuracy \cite{krishnanBenchmarkingEgocentricVisualInertial2025a,camposORBSLAM3AccurateOpenSource2021}.

The field of VIO and VI-SLAM has made rapid progress in the previous decade enabling this technology to be deployed in commercial devices like VR headsets and smartphones. However, existing limitations in terms of accuracy per watt remain in the literature \cite{kumarHighSpeedStereoVisual2024}, driving research into highly specialized hardware accelerators \cite{suleimanNavion2mWFully2019,kuhneLEVIOLightweightEmbedded2026}. These accelerators reach their efficiency through custom silicon, so they only ship in devices designed around them. We instead use a co-processor that existing devices already carry. It is nowadays common to find systems, even low-end ones, that have a GPU alongside the CPU. DSPs have developed from fixed-function audio and modem processors into increasingly general-purpose compute units that run at a fraction of the energy cost of a CPU \cite{codrescuHexagonDSPArchitecture2014}. The Hexagon DSP is now present in a multitude of devices spanning smartphones, VR, AR, and robotics (e.g., Snapdragon 8, XR2, AR1, and RB5 respectively). The DSP provides very long instruction words with wide vector lanes coupled with fast local memory. When properly utilized, this results in significantly reduced power usage for certain tasks compared to using the CPU or GPU. This kind of co-processor allows robots, XR devices, and smartphones to do long-haul vision tasks thanks to the low power required to perform them on these accelerators. The low power not only extends operation, but can also directly improve the target application: smaller batteries for lighter and cheaper drones, lighter headworn devices for reduced neck strain, and cooler smartglasses for improved comfort. The goal we work towards is tracking that runs for a full day on a device's own battery. Even with all these benefits, very few works in the literature exploit it for the fundamental task of visual-inertial tracking and, to the best of our knowledge, none of them for the stereo-inertial sensor modality.

In this work, we offload the frontend of a visual-inertial odometry system based on Basalt \cite{usenkoBasaltVisualInertialMapping2020} to the Hexagon DSP of a Samsung Galaxy S23 (Snapdragon 8 Gen 2, Hexagon V73). A direct port does not beat the CPU. Sparse tracking branches per feature and reads scattered pixels, the vector unit runs all lanes together and its gather reads only on-chip memory, and each DSP call costs hundreds of microseconds, so calling the DSP once per stage spends much of the gain on round trips. Most of the work is restructuring the frontend around these three constraints. The contributions of this work are:

\begin{itemize}
  \item The first stereo-inertial odometry system in the literature leveraging the Hexagon DSP co-processor.
  \item An optimized open-source\footnote{The source code will be released} implementation of the VIO frontend providing insight into the DSP for vision tasks.
  \item A thorough power, latency, and performance analysis that shows HexVIO achieving an 86\% increase of throughput or a 67\% reduction of power (Fig.~\ref{fig:teaser}).
  \item An energy-focused configuration that sustains real-time 30\,fps tracking for 0.83\,W, corresponding to $\sim$18 hours of tracking on the testing device.
\end{itemize}

\section{RELATED WORK}

Contemporary \textbf{VIO} and \textbf{VI-SLAM} pipelines generally decompose into a visual frontend, responsible for keypoint detection and feature tracking, an inertial integration stage, and a backend optimizer or filter tasked with fusing these modalities. A diverse set of representative systems has emerged in the literature, including optimization-based frameworks such as ORB-SLAM3 \cite{camposORBSLAM3AccurateOpenSource2021}, which augments feature-based pipelines with inertial measurements and map atlases, and VINS-Mono/VINS-Fusion \cite{qinVINSMonoRobustVersatile2018} or OKVIS \cite{leuteneggerKeyframebasedVisualInertial2015b}, which rely on sliding-window nonlinear optimization. Filter-based approaches, notably ROVIO \cite{bloeschRobustVisualInertial2015} and MSCKF \cite{mourikisMultiStateConstraintKalman2007a}, also remain widely adopted. In this work, we build upon Basalt \cite{usenkoBasaltVisualInertialMapping2020}, incorporating modifications from \cite{demayoMonadoSLAMDataset2025}. Basalt utilizes a fixed-lag smoother that couples patch-based Lucas-Kanade tracking \cite{Lucas-1981-15101,bouguet2001pyramidalf}, implemented in its inverse-compositional form \cite{Baker-2002-8493} to keep reference-side linear algebra out of the iteration loop, with FAST corner detection \cite{rostenMachineLearningHighSpeed2006}, preintegrated IMU factors, and first-estimate-Jacobian marginalization. The open implementation of Basalt, along with its balance between performance and accuracy \cite{demayoMonadoSLAMDataset2025}, makes it a particularly strong target for frontend acceleration on embedded platforms.

\textbf{Mobile VIO} systems must additionally sustain the camera frame rate within a strict thermal and power envelope, a dual constraint formalized in embedded VIO benchmarking practices that report tracking accuracy alongside per-frame computational cost on target hardware \cite{delmericoBenchmarkComparisonMonocular2018a}. One established strategy to meet these constraints involves redesigning the estimator itself. For instance, LEVIO \cite{kuhneLEVIOLightweightEmbedded2026} achieves 20\,fps six-DoF tracking below 100\,mW on an ultra-low-power RISC-V SoC by employing a streamlined ORB-based frontend alongside a comprehensive hardware-software co-design. In contrast, this work preserves Basalt's complete pipeline and instead focuses on accelerating its execution. This is a reasonable approach given these SoCs are already widely adopted by different industries trying to leverage spatial computing.

The visual frontend is a natural \textbf{hardware acceleration} target, as its per-feature computations are easy to parallelize and its backend interface reduces to sparse correspondences. Prior work has largely targeted GPUs: vilib \cite{nagyFasterFASTGPUAccelerated2020a} and Jetson-SLAM \cite{kumarHighSpeedStereoVisual2024} accelerate FAST-based frontends in CUDA (KLT as in ours, and ORB, respectively), while FastTrack \cite{khabiriFastTrackGPUAcceleratedTracking2025a} offloads portions of ORB-SLAM3 to the GPU. GLidE-SLAM \cite{desousaGLidESLAMGLAcceleratedIndirectDirect2026a} employs OpenGL compute shaders to circumvent CUDA lock-in and augments indirect features with direct GPU tracking. In contrast, our DSP-based approach frees the GPU for concurrent workloads and sidesteps the thermal constraints and higher power draw typical of GPU acceleration, while still delivering low-power, high-frame-rate tracking.

At the other extreme, \textbf{custom silicon} achieves the lowest reported energy per frame. Navion \cite{suleimanNavion2mWFully2019} integrates a complete VIO pipeline into a 2\,mW ASIC via algorithm-hardware co-design \cite{zhangVisualInertialOdometryChip2017}, albeit at the cost of fixing the algorithm. FPGA-based solutions offer a more flexible alternative: eSLAM \cite{liuESLAMEnergyEfficientAccelerator2019} accelerates only ORB extraction and matching, while ac2SLAM \cite{wangAc2SLAMFPGAAccelerated2021} accelerates both the ORB-SLAM2 frontend and backend, though neither incorporates inertial fusion. PX4Flow \cite{honeggerOpenSourceOpen2013} instead computes block-matching optical flow on a microcontroller inside the sensor. Compared to the Hexagon DSP, however, these custom accelerators lack broad availability on commodity platforms, and ASICs offer no post-deployment reprogrammability.

\textbf{Hexagon DSPs} occupy a middle ground: they are programmable, widely deployed, and optimized for energy-efficient single-instruction multiple-data (SIMD) vector processing. The Hexagon architecture \cite{codrescuHexagonDSPArchitecture2014} combines a VLIW scalar core with the HVX wide-vector coprocessor and a software-managed scratchpad. Originally designed for audio and modem tasks, it now supports general-purpose workloads, including LLM inference \cite{haoScalingLLMTestTime2026}, while vendor CV/ML stacks (FastCV, SNPE, QNN \cite{qualcomm2024qnni}) and the Halide compiler \cite{ragan-kelleyHalideLanguageCompiler2013} already target it for dense image processing. Despite this ecosystem, academic use of Hexagon for visual-inertial tracking remains limited. The closest prior work is Muffang's thesis \cite{muffang2021slamh}, which offloaded OpenVINS \cite{genevaOpenVINSResearchPlatform2020a} frontend routines to a Snapdragon 845 via FastCV; it reduced CPU load but the offloaded kernels ran slower than their CPU equivalents, and no source code was released. An open Hexagon port of the ORB-SLAM3 feature extractor and matcher exists for the Lynx R1 headset \cite{lynx2025orbslam3g}, but it lacks inertial fusion and requires a rooted device. Crucially, none of the prior efforts support stereo-inertial tracking, a ubiquitous modality in consumer XR and robotics.

Relative to this body of work, our system is, to the best of our knowledge, the \textit{first} to accelerate a multicamera VIO pipeline on the Hexagon DSP of a stock consumer smartphone. While LEVIO shares our energy efficiency goals, it fundamentally redesigns the estimator for minimal compute budgets sacrificing accuracy; approaches like vilib, FastTrack, Jetson-SLAM, and GLidE-SLAM target frontend acceleration but rely on higher-power GPUs; ASIC-, FPGA- and MCU-based solutions like Navion, eSLAM, ac2SLAM, and PX4Flow achieve high efficiency but require hardware that is unavailable on common platforms. Prior Hexagon-based tracking efforts either lack stereo-inertial support, remain closed-source, and/or require rooted devices. In contrast, our implementation is open-source, runs on unmodified consumer hardware, and demonstrates that the Hexagon DSP can effectively accelerate stereo-inertial VIO frontends at significantly lower power.

\section{METHOD}
\label{sec:method}

\subsection{Architecture}
\label{sec:architecture}
Two properties of the V73 shape the frontend, its vector execution model and its memory hierarchy.

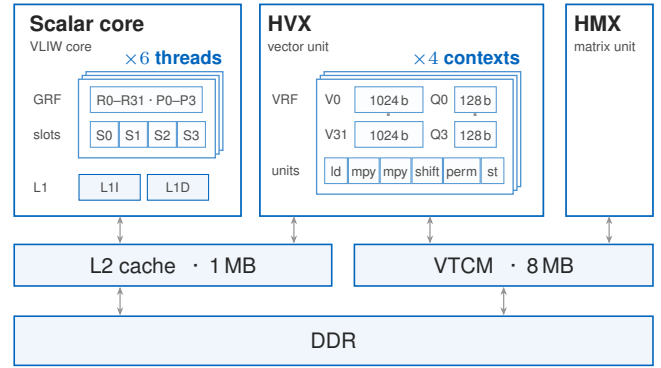
\begin{figure}[tbp]
  \centering
  \definecolor{tumblue}{HTML}{0065BD}

\begin{tikzpicture}[x=1mm, y=1mm,
    font=\sffamily\footnotesize,
    compute/.style={rectangle, draw=tumblue, line width=.65pt, fill=white, inner sep=0pt},
    ctx/.style={rectangle, draw=tumblue!70, line width=.5pt, fill=white, inner sep=0pt},
    resource/.style={rectangle, draw=tumblue!65, line width=.4pt, fill=tumblue!1,
        minimum height=3.4mm, inner sep=.4mm, font=\sffamily\tiny, text=black!70,
        text height=1.1ex, text depth=.25ex, align=center},
    memory/.style={rectangle, draw=tumblue, line width=.60pt, fill=tumblue!6,
        inner sep=.8mm, align=center, text=black!85},
    memcell/.style={resource, fill=tumblue!6, draw=tumblue},
    head/.style={font=\sffamily\footnotesize\bfseries, text=black!85, inner sep=0pt},
    desc/.style={font=\sffamily\tiny, text=black!70, inner sep=0pt, align=left},
    count/.style={font=\sffamily\scriptsize\bfseries, text=tumblue, inner sep=0pt},
    >={Stealth[length=1.1mm, width=.85mm]},
    flow/.style={<->, draw=black!42, line width=.45pt},
]

\coordinate (sSW) at (2,21.5);    \coordinate (sNE) at (32,49.5);
\coordinate (hSW) at (34.5,21.5); \coordinate (hNE) at (72.0,49.5);
\coordinate (mSW) at (75.0,21.5); \coordinate (mNE) at (86.5,49.5);
\coordinate (tSW) at (10.5,29.2); \coordinate (tNE) at (28.6,39.6);
\coordinate (cSW) at (42.0,24.4);  \coordinate (cNE) at (68.0,39.6);

\begin{pgfonlayer}{background}
\node[compute, fit=(sSW)(sNE)] (scalar) {};
\node[compute, fit=(hSW)(hNE)] (hvx) {};
\node[compute, fit=(mSW)(mNE)] (hmx) {};
\foreach \d in {1.0,0.5} {
  \filldraw[fill=white, draw=tumblue!70, line width=.5pt]
    ($(tSW)+(\d,\d)$) rectangle ($(tNE)+(\d,\d)$);
  \filldraw[fill=white, draw=tumblue!70, line width=.5pt]
    ($(cSW)+(\d,\d)$) rectangle ($(cNE)+(\d,\d)$);
}
\node[ctx, fit=(tSW)(tNE)] (thread) {};
\node[ctx, fit=(cSW)(cNE)] (vctx) {};
\end{pgfonlayer}

\node[head, anchor=west] at (4.0,47.0) {Scalar core};
\node[desc, anchor=west] at (4.0,44.0) {VLIW core};
\node[count, anchor=south east] at (29.6,41.2) {$\times 6$ threads};
\node[desc, anchor=west] at (4.5,36.9) {GRF};
\node[resource, minimum width=14.8mm, anchor=west] (grf) at (12.0,36.9)
    {R0--R31 \textperiodcentered\ P0--P3};
\node[desc, anchor=west] at (4.5,32.3) {slots};
\node[resource, minimum width=3.7mm, anchor=west] (s0) at (12.0,32.3) {S0};
\node[resource, minimum width=3.7mm, anchor=west] (s1) at (s0.east) {S1};
\node[resource, minimum width=3.7mm, anchor=west] (s2) at (s1.east) {S2};
\node[resource, minimum width=3.7mm, anchor=west] (s3) at (s2.east) {S3};
\node[desc, anchor=west] at (4.5,25.3) {L1};
\node[memcell, minimum width=8.1mm, anchor=west] (l1i) at (10.5,25.3) {L1I};
\node[memcell, minimum width=8.1mm, anchor=west] (l1d) at ($(l1i.east)+(0.8,0)$) {L1D};

\node[head, anchor=west] at (35.5,47.0) {HVX};
\node[desc, anchor=west] at (35.5,44.0) {vector unit};
\node[count, anchor=south east] at (69.0,41.2) {$\times 4$ contexts};
\node[desc, anchor=west] at (36.0,36.9) {VRF};
\node[desc, anchor=west] at (43.0,36.9) {V0};
\node[resource, minimum width=9.0mm, anchor=west] (vr0) at (47.0,36.9) {1024\,b};
\node[desc, anchor=west] at (57.0,36.9) {Q0};
\node[resource, minimum width=5.5mm, anchor=west] (q0) at (60.2,36.9) {128\,b};
\node[font=\sffamily\normalsize, text=black!45, inner sep=0pt] at (51.5,34.6) {$\vdots$};
\node[font=\sffamily\normalsize, text=black!45, inner sep=0pt] at (63.0,34.6) {$\vdots$};
\node[desc, anchor=west] at (43.0,32.3) {V31};
\node[resource, minimum width=9.0mm, anchor=west] (vr31) at (47.0,32.3) {1024\,b};
\node[desc, anchor=west] at (57.0,32.3) {Q3};
\node[resource, minimum width=5.5mm, anchor=west] (q3) at (60.2,32.3) {128\,b};
\node[desc, anchor=west] at (36.0,27.5) {units};
\node[resource, minimum width=3.0mm, anchor=west] (r1) at (43.0,27.5) {ld};
\node[resource, minimum width=3.7mm, anchor=west] (r2) at (r1.east) {mpy};
\node[resource, minimum width=3.7mm, anchor=west] (r3) at (r2.east) {mpy};
\node[resource, minimum width=3.7mm, anchor=west] (r4) at (r3.east) {shift};
\node[resource, minimum width=3.9mm, anchor=west] (r5) at (r4.east) {perm};
\node[resource, minimum width=3.0mm, anchor=west] (r6) at (r5.east) {st};

\node[head, anchor=west] at (76.0,47.0) {HMX};
\node[desc, anchor=west] at (76.0,44.0) {matrix unit};

\node[memory, minimum width=42mm, minimum height=5.5mm] (l2) at (23,15)
    {L2 cache \enspace\textperiodcentered\enspace 1\,MB};
\node[memory, minimum width=39.5mm, minimum height=5.5mm] (vtcm) at (66.75,15)
    {VTCM \enspace\textperiodcentered\enspace 8\,MB};
\node[memory, minimum width=84.5mm, minimum height=6.5mm]
    (ddr) at (44.25,5) {DDR};

\draw[flow] (16,21.5) -- (l2.north -| 16,0);
\draw[flow] (39,21.5) -- (l2.north -| 39,0);
\draw[flow] (57,21.5) -- (vtcm.north -| 57,0);
\draw[flow] (l2.south -| 16,0) -- (ddr.north -| 16,0);
\draw[flow] (vtcm.south -| 66.75,0) -- (ddr.north -| 66.75,0);
\draw[flow] (80.7,21.5) -- (vtcm.north -| 80.7,0);

\end{tikzpicture}
  \caption{\textbf{The Hexagon V73 compute DSP.} The scalar core, HVX vector unit and HMX matrix unit share one memory hierarchy. Each hardware thread has its own scalar registers (GRF, predicates P0--P3); each vector context has its own vector registers (VRF, predicates Q0--Q3) and the six functional units that operate on them. The gather instruction, the basis of sparse feature sampling, addresses VTCM only.}
  \label{fig:v73}
\end{figure}

\paragraph*{Computation units}
The Hexagon V73 compute DSP (cDSP) comprises a scalar core and two coprocessors (Fig.~\ref{fig:v73}): HVX (Hexagon Vector eXtensions) for wide SIMD on 1024-bit registers, which we use, and HMX for matrix operations. Every vector instruction occupies one of the four VLIW slots in a thread's instruction packet, so scalar and vector instructions issue together, subject to the available vector functional units. The six hardware threads attach dynamically to the four vector contexts, and peak vector throughput requires all four active.

\paragraph*{Memory subsystem}
The CPU invokes code on the DSP through FastRPC, a remote-procedure-call layer~\cite{qualcomm_fastrpc}. We allocate every buffer that crosses the boundary as shared memory, which FastRPC maps into the DSP's address space instead of copying it. On the DSP side, HVX bypasses the scalar core's L1 and reads from L2 or from VTCM (vector tightly coupled memory), an 8\,MB scratchpad under software control. Alongside ordinary loads and stores, DMA can move bulk data from DDR into VTCM. The vector gather instruction can only address VTCM~\cite[Sec.~3.3]{qualcomm_v73_hvx_prm}, so anything sampled at non-contiguous memory addresses must be resident there rather than merely cached.

\begin{figure}[tbp]
  \centering
  \definecolor{tumblue}{HTML}{0065BD}

\begin{tikzpicture}[x=1mm, y=1mm,
    font=\sffamily\footnotesize,
    panel/.style={rectangle, draw=tumblue!70, line width=.55pt, fill=white, inner sep=0pt},
    px/.style={rectangle, draw=black!35, line width=.3pt, fill=white,
        minimum width=2.6mm, minimum height=2.6mm, inner sep=0pt},
    pxon/.style={px, fill=tumblue!18, draw=tumblue},
    lane/.style={rectangle, draw=tumblue, line width=.4pt, fill=tumblue!6,
        minimum width=2.6mm, minimum height=3.2mm, inner sep=0pt,
        font=\sffamily\tiny, text=black!70, text height=1.1ex, text depth=.25ex},
    feat/.style={circle, draw=tumblue, line width=.4pt, fill=tumblue!18,
        minimum size=2.2mm, inner sep=0pt},
    head/.style={font=\sffamily\footnotesize\bfseries, text=black!85, inner sep=0pt},
    desc/.style={font=\sffamily\tiny, text=black!55, inner sep=0pt, align=left},
    axis/.style={font=\sffamily\scriptsize, text=black!80, inner sep=0pt, align=left},
    note/.style={font=\sffamily\tiny, text=black!70, inner sep=0pt, align=center},
    >={Stealth[length=1.1mm, width=.85mm]},
    flow/.style={->, draw=black!42, line width=.45pt},
]

\coordinate (lSW) at (0,5);  \coordinate (lNE) at (41,36.3);
\begin{pgfonlayer}{background}
\node[panel, fit=(lSW)(lNE)] {};
\end{pgfonlayer}
\node[head, anchor=west] at (2,33.8) {Pixel-parallel};
\node[desc, anchor=east] at (39,33.8) {pyramid, FAST};

\foreach \r in {0,...,6}
  \foreach \c in {0,...,10} {
    \ifnum\r=2
      \node[pxon] at (4.4+\c*2.7, 29.9-\r*2.7) {};
    \else
      \node[px] at (4.4+\c*2.7, 29.9-\r*2.7) {};
    \fi}
\foreach \dx/\dy in {0/-3,1/-3,2/-2,3/-1,3/0,3/1,2/2,1/3,0/3,-1/3,-2/2,-3/1,-3/0,-3/-1,-2/-2,-1/-3}
  \node[circle, draw=tumblue, line width=.35pt, minimum size=1.3mm, inner sep=0pt]
    at ({4.4+(5+\dx)*2.7}, {29.9-(3+\dy)*2.7}) {};
\node[circle, fill=tumblue, minimum size=1.3mm, inner sep=0pt] (cp) at (4.4+5*2.7, 29.9-3*2.7) {};

\foreach \i in {0,...,10}
  \node[lane] at (4.4+\i*2.7, 8.4) {};
\draw[->, draw=black!22, line width=.35pt]
  ({4.4+(4+3)*2.7}, {29.9-2*2.7-1.3}) -- ({4.4+4*2.7}, {8.4+1.6});
\draw[->, draw=tumblue, line width=.45pt]
  ({4.4+(5+3)*2.7}, {29.9-2*2.7-1.3}) -- ({4.4+5*2.7}, {8.4+1.6});

\coordinate (rSW) at (44,5); \coordinate (rNE) at (85,36.3);
\begin{pgfonlayer}{background}
\node[panel, fit=(rSW)(rNE)] {};
\end{pgfonlayer}
\node[head, anchor=west] at (46,33.8) {Feature-parallel};
\node[desc, anchor=east] at (83,33.8) {tracking};

\foreach \r in {0,...,6}
  \foreach \c in {0,...,10}
    \node[px] at (48.4+\c*2.7, 29.9-\r*2.7) {};
\foreach \cx/\cy/\sx/\sy/\l/\yk in {1.6/1.4/3/2/0/13.6, 5.6/3.6/7/4/1/12.4, 8.0/1.1/9/2/2/11.2} {
  \foreach \dc in {0,1} \foreach \dr in {0,1}
    \node[px, fill=tumblue!22, draw=tumblue!70] at ({48.4+(\sx+\dc)*2.7}, {29.9-(\sy+\dr)*2.7}) {};
  \node[circle, draw=tumblue, line width=.45pt, fill=white, minimum size=1.7mm, inner sep=0pt]
    at ({48.4+\cx*2.7}, {29.9-\cy*2.7}) {};
  \node[note, text=tumblue, anchor=south east, inner sep=.2mm] at ({48.4+\cx*2.7-0.6}, {29.9-\cy*2.7+0.6}) {\l};
  \node[circle, fill=tumblue, minimum size=1.2mm, inner sep=0pt]
    at ({48.4+(\cx+1.7)*2.7}, {29.9-(\cy+1.3)*2.7}) {};
  \draw[->, draw=tumblue!60, line width=.35pt, shorten >=.3mm, shorten <=.9mm]
    ({48.4+\cx*2.7}, {29.9-\cy*2.7}) -- ({48.4+(\cx+1.7)*2.7}, {29.9-(\cy+1.3)*2.7});
  \draw[->, draw=tumblue, line width=.45pt]
    ({48.4+(\sx+0.5)*2.7}, {29.9-(\sy+1)*2.7-1.35}) -- ({48.4+(\sx+0.5)*2.7}, \yk) -- ({48.4+\l*2.7}, \yk) -- ({48.4+\l*2.7}, {8.4+1.6});
  \node[lane, fill=tumblue!22, draw=tumblue] at (48.4+\l*2.7, 8.4) {\l};
}
\foreach \i in {3,...,10}
  \node[lane] at (48.4+\i*2.7, 8.4) {};

\end{tikzpicture}
  \caption{\textbf{Two vectorization axes.} Image stages use one pixel per lane: adjacent
  pixels map to adjacent lanes, and for FAST, ring point $k$ for all 128
  candidates is row $y{+}dy_k$ shifted by $dx_k$ lanes. Tracking uses
  one feature per lane: at each offset of the shared 52-point pattern,
  every lane gathers the word holding its own feature's pixel pair, so 32
  features advance in lockstep.}
  \label{fig:lanes}
\end{figure}
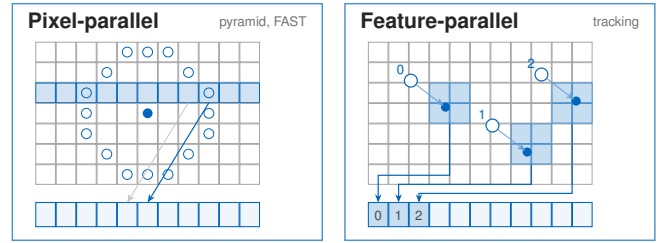

\begin{figure}[tbp]
  \centering
  \definecolor{tumblue}{HTML}{0065BD}

\begin{tikzpicture}[x=1mm, y=1mm,
    font=\sffamily\footnotesize,
    host/.style={rectangle, draw=black!55, line width=.65pt, fill=white,
        inner sep=0pt, align=center, font=\sffamily\scriptsize, text=black!85},
    stage/.style={rectangle, draw=tumblue!70, line width=.5pt, fill=white,
        inner sep=0pt, align=center, font=\sffamily\scriptsize, text=black!80,
        minimum width=14mm, minimum height=8mm},
    memory/.style={rectangle, draw=tumblue, line width=.60pt, fill=tumblue!6,
        inner sep=0pt, align=center, font=\sffamily\scriptsize, text=black!85},
    fused/.style={rectangle, draw=tumblue!45, line width=.5pt, fill=tumblue!3, inner sep=0pt},
    pill/.style={rectangle, rounded corners=2.6mm, draw=black!50, line width=.45pt,
        fill=white, inner sep=0pt, align=center, font=\sffamily\tiny, text=black!75,
        minimum height=5.2mm},
    desc/.style={font=\sffamily\tiny, text=black!55, inner sep=0pt, align=left},
    side/.style={font=\sffamily\scriptsize\bfseries, inner sep=0pt},
    >={Stealth[length=1.1mm, width=.85mm]},
    flow/.style={->, draw=black!42, line width=.45pt},
    bulk/.style={->, draw=black!45, line width=1.1pt},
    rpc/.style={->, draw=tumblue, line width=.5pt},
]

\node[host, minimum width=25mm, minimum height=8mm] (ddr) at (13,51.5)
    {Camera driver / replay\\{\tiny\color{black!55}shared DDR buffer}};
\node[host, minimum width=25mm, minimum height=8mm] (disp) at (40,51.5)
    {Frontend dispatch};
\node[host, minimum width=22mm, minimum height=8mm] (be) at (76,51.5)
    {Backend estimator\\{\tiny\color{black!55}IMU preintegration}};

\draw[flow] ([yshift=1.5mm]disp.east) -- ([yshift=1.5mm]be.west);
\node[desc, anchor=south] at (58.75,56) {observations};
\draw[flow] ([yshift=-1.5mm]be.west) -- ([yshift=-1.5mm]disp.east);
\node[desc, anchor=north] at (58.75,47) {pose, depth};

\draw[densely dotted, draw=black!75, line width=.7pt] (0,40) -- (88,40);
\node[side, text=black!60, anchor=south east] at (86.7,40.7) {CPU};
\node[side, text=tumblue, anchor=north east] at (86.7,39.3) {cDSP};

\node[pill, minimum width=14mm] (dimg) at (13,40) {stereo pair};
\node[pill, minimum width=17mm] (dcall) at (32,40) {point set, pose\\seed, depth seed};
\node[pill, minimum width=15mm] (dret) at (50,40) {tracks, warps,\\new corners};

\coordinate (fSW) at (5.5,16.3);  \coordinate (fNE) at (86.7,30.2);
\begin{pgfonlayer}{background}
\node[fused, fit=(fSW)(fNE)] (fusedbox) {};
\end{pgfonlayer}
\node[desc, text=tumblue!80, anchor=north west] at ([xshift=1.2mm, yshift=-1mm]fusedbox.north west) {fused call};

\node[stage] (s1) at (13.9,22.5) {Pyramid\\build};
\node[stage] (s2) at (30.0,22.5) {Temporal\\tracking};
\node[stage] (s3) at (46.1,22.5) {Detection L};
\node[stage] (s4) at (62.2,22.5) {Stereo\\tracking};
\node[stage] (s5) at (78.3,22.5) {Detection R};
\draw[flow] (s1) -- (s2); \draw[flow] (s2) -- (s3);
\draw[flow] (s3) -- (s4); \draw[flow] (s4) -- (s5);

\node[memory, minimum width=81.2mm, minimum height=7mm] (vtcm) at (46.1,8)
    {VTCM \enspace\textperiodcentered\enspace four pyramid slots:
     current / previous $\times$ left / right};
\foreach \s in {s1,s2,s3,s4,s5}
  \draw[flow, <->] (\s.south) -- (\s.south |- vtcm.north);

\draw[bulk] (ddr.south) -- (dimg.north);
\draw[bulk] (dimg.south) -- ++(0,-2) -| (3,8) -- (vtcm.west);
\node[desc, anchor=west] at (3.8,34.5) {DMA};
\draw[rpc]  (disp.south -| dcall.north) -- (dcall.north);
\draw[rpc]  (dcall.south) -- (dcall.south |- fusedbox.north);
\draw[rpc]  (dret.north |- fusedbox.north) -- (dret.south);
\draw[rpc]  (dret.north) -- (disp.south -| dret.north);

\end{tikzpicture}
  \caption{\textbf{System partition.} The frontend runs on the cDSP in one fused FastRPC call per stereo frame; inertial integration and the estimator stay on the CPU. The call fetches the full-resolution stereo pair from a shared DDR buffer into VTCM by DMA; every other argument is sparse feature data, passed by shared-memory mapping rather than copied. The current and previous stereo pyramids remain resident in VTCM.}
  \label{fig:system}
\end{figure}
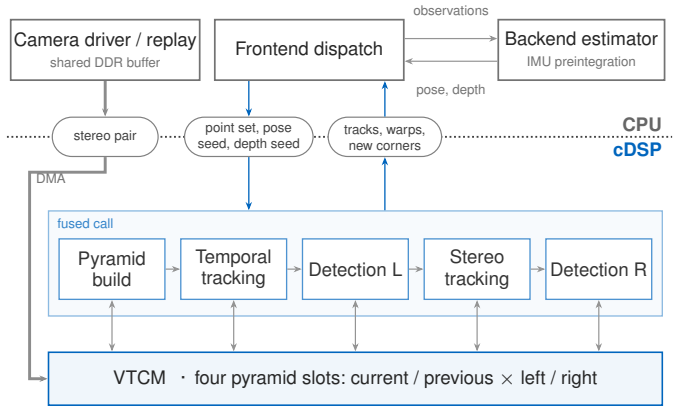

\subsection{Pipeline}
\label{sec:pipeline}

Basalt's frontend has three stages, pyramid build, FAST-9 detection and sparse inverse-compositional KLT tracking (Fig.~\ref{fig:system}). We run all three on the DSP and fuse them into one FastRPC call per frame. Each call has a fixed cost. One that carries the image buffers but does no work takes about 0.25\,ms at the DSP's top clock and 0.35\,ms at its low-voltage corner under the polling policy, and 0.65\,ms under the driver defaults, which let the CPU sleep while it waits and the DSP idle between calls. One call per stage and camera, plus one for stereo matching, would mean seven calls per frame; fusing them amortizes the fixed cost over the whole frontend (Fig.~\ref{fig:fusion}).

Apart from the images, the call's inputs and outputs are sparse feature data that scale with the feature count rather than the image size. The image pyramids must reside in VTCM for the tracking gathers, so the call begins by fetching the full-resolution stereo pair from the shared buffer by DMA and builds the coarser levels on the DSP. A coarse occupancy grid links the stages. Surviving temporal tracks prefill it, so detection searches only cells that still need features; the new left-camera detections then seed stereo tracking, whose matches likewise occupy cells before right-camera detection. Right-camera detection also skips cells that project into the left camera's view, as in \cite{demayoMonadoSLAMDataset2025}.

The frontend uses two vectorization strategies (Fig.~\ref{fig:lanes}). Pyramid construction and detection are pixel-parallel. Neighboring pixels are contiguous in memory, so one vector load brings in 64 16-bit pyramid pixels or 128 8-bit detector pixels, one per lane, and each lane computes its pixel's part of the stencil or ring test. Tracking is feature-parallel. Each feature has a KLT patch of 52 sample points spread around it, so a lane holds one feature instead. The 32 features of a batch move through the 52 points in lockstep, and at each point a gather fetches the pixels each feature needs to interpolate it.

\subsection{Pyramid}
\label{sec:pyramid}

To perform pyramidal KLT tracking \cite{bouguet2001pyramidalf}, we build a four-level image pyramid per camera. Each level is a separable $5\times5$ binomial blur ($w = [1,4,6,4,1]$, with $16^2 = 256$ normalization) followed by $2{:}1$ subsampling. The vertical pass writes a 32-bit temporary $V_\ell$, since each pass can grow the range by $16\times$. Division by 256 with rounding becomes an addition of $128$ and a right shift by eight. Let $\mathbf V_\ell[r,:]$ be row $r$ of the temporary, one column per lane; the horizontal pass computes output lane $n$ of row $r$ as

\begin{align}
\label{eq:pyrvector}
L_{\ell+1}[r,n] = \Big(\sum_{i=-2}^{2} w_i\, V_\ell[r,2n + i] + 128\Big) \gg 8.
\end{align}

The horizontal pass first deinterleaves the row of $\mathbf V_\ell$ into its even and odd columns with one vector deal, so the filter is evaluated only at the columns the $2{:}1$ subsampling keeps, and the five terms then need only three one-lane alignments; the filter itself needs no other cross-lane traffic.

One VTCM slot per camera holds all four levels at fixed offsets, with rows padded to a multiple of 128\,B so every row load is aligned. Levels 1--3 are built directly in VTCM, and nothing is written back to DDR; apart from the host's copy into the shared buffer, the level-0 DMA is the frontend's only image-sized DDR traffic. Four workers (hardware threads) build the pyramids, two per camera, each owning a disjoint band of rows and synchronizing once between levels.

\subsection{Detection}
\label{sec:detection}

We detect FAST-9 corners \cite{rostenMachineLearningHighSpeed2006}. A pixel $p$ is a corner if nine contiguous pixels on the radius-3 Bresenham circle are all brighter than $I(p)+\tau$ or all darker than $I(p)-\tau$, and its score is the largest $\tau$ that satisfies this condition. Basalt keeps at most one corner per $50\times50$ cell by running thresholds $\langle 40,20,10,5\rangle$ from strict to lenient and taking the highest-response candidate at the first threshold that yields one. Since the response is the largest passing threshold, that procedure returns the highest-scoring corner in the cell above the minimum threshold. We therefore make one full-image pass at the lowest threshold, $\tau=5$, computing the corner test and the score over the ring, and keep the per-cell maximum after non-maximum suppression, obtaining in one vectorized sweep the same corners as Basalt.

With an 8-bit working image in VTCM, 128 adjacent candidates of a row form one \textit{vector block} (Fig.~\ref{fig:lanes}). Ring point $k$ has a fixed offset $(dx_k, dy_k)$ from its candidate with $|dy_k|\le 3$, so for all 128 candidates it is row $y+dy_k$ shifted by $dx_k$ lanes, and one vector comparison against $I(p)+\tau$ tests it for every candidate at once, yielding a lane mask $\mathbf m^{+}_k$. The comparison against $I(p)-\tau$ gives $\mathbf m^{-}_k$. Nine contiguous bright points exist if nine consecutive masks are set for some start $k$, wrapping around the ring:

\begin{equation}
  \mathbf c^{+} = \bigvee_{k=0}^{15}\;\bigwedge_{m=0}^{8}\;
  \mathbf m^{+}_{(k+m)\bmod 16},
  \label{eq:fast-lane}
\end{equation}

$\mathbf c^{-}$ follows from the dark masks, a lane is a corner if $\mathbf c^{+}\vee\mathbf c^{-}$, and the score is the maximum, over all arc starts, of the minimum brightness difference along the arc. An occupancy bitset over the cells, folded into one mask per cell row, lets a worker skip a whole row when all its cells are occupied, or skip a vector block with a single mask test when every cell it spans is occupied. Four workers partition cell rows and merge their private results on completion.

\subsection{Tracking}
\label{sec:tracking}

We track features with patch-based inverse-compositional KLT \cite{Lucas-1981-15101,bouguet2001pyramidalf}. Each feature carries an $SE(2)$ warp $\mathbf T(\boldsymbol\xi)$, $\boldsymbol\xi=(t_x,t_y,\theta)\in\mathbb R^3$, estimated by minimizing the mean-normalized SSD over a fixed $52$-point pattern $\Omega=\{\mathbf x_i\}$ centered on the feature \eqref{eq:track-residual}, where $\bar I$ is the mean pattern intensity,
\begin{equation}
  e_i(\boldsymbol\xi)=\frac{I_{t+1}\big(\mathbf T(\boldsymbol\xi)\,\mathbf
  x_i\big)}{\bar I_{t+1}}-\frac{I_t(\mathbf x_i)}{\bar I_t},
  \qquad \mathbf x_i\in\Omega.
  \label{eq:track-residual}
\end{equation}
Each iteration applies $\boldsymbol\delta=-G\,\mathbf e$ as $\mathbf T\leftarrow\mathbf T\circ\exp(\boldsymbol\delta)$. A tracking call seeds each feature by reprojecting it with the predicted pose and depth. It then builds the feature's reference once per pyramid level, comprising the samples, gradients, the Jacobian rows $J_i=[\,g_x,\ g_y,\ g_y u_i-g_x v_i\,]$ at $\mathbf x_i=(u_i,v_i)$, the Hessian $H=\sum_i J_i^{\top}J_i$ and the pseudoinverse $G=H^{-1}J^{\top}$. Five coarse-to-fine Gauss-Newton iterations per level re-sample only the target and apply the precomputed $G$. A forward-backward check then discards tracks that do not return. The per-feature matrices are too small to benefit from HMX.

Basalt samples each pattern point in a scalar loop and vectorizes only the sums that follow. We invert this, with one feature per lane, 32 per batch, the pattern index as the sequential loop (Fig.~\ref{fig:lanes}), so sampling is vectorized across features and no reduction crosses lanes. By stacking a batch column-wise, we transform each per-feature vector into a $52\times32$ matrix, specifically the residuals $\mathbf E$ and reference samples $\mathbf R$, while the per-feature $3\times52$ pseudoinverses are represented collectively by three such matrices $\mathbf G^{(x)},\mathbf G^{(y)},\mathbf G^{(\theta)}$. In this layout, each row is a single $32$-lane vector, allowing the update \eqref{eq:track-lane} to be computed as 52 rows of three multiply-adds. The Hessian is computed via six element-wise accumulations and its inverse via a pivoted $\mathrm{LDL}^{\top}$ with predicated swaps; the 32 inversions run without branching. Invalid features are rejected by lane predication rather than branches, and a batch exits only once all 32 lanes have failed.

\begin{equation}
  \boldsymbol\Delta=-\sum_{i=1}^{52}
  \begin{bmatrix}\mathbf G^{(x)}_{i,:}\\[1pt] \mathbf G^{(y)}_{i,:}\\[1pt]
  \mathbf G^{(\theta)}_{i,:}\end{bmatrix}\odot\mathbf E_{i,:}\;\in\;
  \mathbb R^{3\times32}.
  \label{eq:track-lane}
\end{equation}

The inner loop reads two kinds of data. The reference samples $\mathbf R$ and the three pseudoinverse rows do not change between iterations, so they are stored row by row in their $52\times32$ form, and a pattern point costs four aligned loads with no cross-lane traffic. The target samples change every iteration and are re-fetched by predicated vector gathers (Sec.~\ref{sec:architecture}), two per pattern point. Each 32-bit gather retrieves two adjacent 16-bit pixels of one row, so two gathers supply the $2\times2$ bilinear neighborhood, with invalid features masked in the instruction. We measure a gather latency of about 60 cycles on this device, so we issue the gathers for point $i$ while interpolating point $i{-}1$ and avoid stalling on the results. Finally, the backward pass compacts survivors into dense batches and rebuilds only their reference patches, so patch build and solve run only for surviving features.

\section{RESULTS}
\label{sec:results}

\begin{table*}[t]
    \centering
    \caption{Maximum throughput runs: accuracy and times of CPU baseline vs. HexVIO (DSP)}
    \label{tab:performance}
    \resizebox{\textwidth}{!}{%
    \setlength{\tabcolsep}{3.5pt}
    \begin{tabular}{l *{14}{c}}

        \toprule
        & \multicolumn{2}{c}{\textbf{ATE [cm]}}
        & \multicolumn{2}{c}{\textbf{RTE [cm]}}
        & \multicolumn{2}{c}{\textbf{FPS}}
        & \multicolumn{2}{c}{\textbf{Frontend [ms]}}
        & \multicolumn{2}{c}{\textbf{Backend [ms]}}
        & \multicolumn{2}{c}{\textbf{End-to-end [ms]}}
        & \multicolumn{2}{c}{\textbf{p99 [ms]}}
        \\
        
        \cmidrule(lr){2-3}
        \cmidrule(lr){4-5}
        \cmidrule(lr){6-7}
        \cmidrule(lr){8-9}
        \cmidrule(lr){10-11}
        \cmidrule(lr){12-13}
        \cmidrule(lr){14-15}
        
        \textbf{Sequence}
        & CPU & DSP
        & CPU & DSP
        & CPU & DSP
        & CPU & DSP
        & CPU & DSP
        & CPU & DSP
        & CPU & DSP
        \\
        
        \midrule
        
        EV102
        & $5.40$ & $5.41$
        & $1.12$ & $1.12$
        & $70.1$ & $\mathbf{120.0}$
        & $7.75 \pm 2.25$ & $\mathbf{2.32 \pm 0.29}$
        & $7.15 \pm 2.87$ & $\mathbf{5.98 \pm 2.30}$
        & $15.00 \pm 3.12$ & $\mathbf{8.38 \pm 2.48}$
        & $23.5$ & $\mathbf{14.7}$
        \\

        EMH03
        & $6.05$ & $6.43$
        & $1.18$ & $1.18$
        & $56.2$ & $\mathbf{94.9}$
        & $9.05 \pm 1.81$ & $\mathbf{2.47 \pm 0.32}$
        & $9.32 \pm 4.32$ & $\mathbf{7.97 \pm 3.42}$
        & $18.48 \pm 4.57$ & $\mathbf{10.52 \pm 3.63}$
        & $31.3$ & $\mathbf{19.2}$
        \\

        TR2
        & $7.30$ & $7.33$
        & $0.70$ & $0.70$
        & $105.6$ & $\mathbf{180.1}$
        & $5.79 \pm 1.20$ & $\mathbf{2.00 \pm 0.22}$
        & $3.88 \pm 1.56$ & $\mathbf{3.55 \pm 1.48}$
        & $9.59 \pm 1.90$ & $\mathbf{5.62 \pm 1.64}$
        & $15.6$ & $\mathbf{10.3}$
        \\

        TR6
        & $2.06$ & $2.00$
        & $0.38$ & $0.38$
        & $93.7$ & $\mathbf{155.0}$
        & $6.54 \pm 1.61$ & $\mathbf{2.18 \pm 0.24}$
        & $4.65 \pm 1.71$ & $\mathbf{4.30 \pm 1.50}$
        & $10.87 \pm 2.19$ & $\mathbf{6.55 \pm 1.64}$
        & $17.4$ & $\mathbf{11.4}$
        \\

        MOO02
        & $23.48$ & $23.43$
        & $1.54$ & $1.54$
        & $103.2$ & $\mathbf{206.3}$
        & $6.68 \pm 0.73$ & $\mathbf{2.10 \pm 0.13}$
        & $2.80 \pm 1.81$ & $\mathbf{2.51 \pm 1.54}$
        & $9.68 \pm 1.98$ & $\mathbf{4.78 \pm 1.62}$
        & $16.0$ & $\mathbf{10.4}$
        \\

        MOO08
        & $2.74$ & $2.73$
        & $1.12$ & $1.11$
        & $97.6$ & $\mathbf{177.1}$
        & $6.83 \pm 1.37$ & $\mathbf{2.14 \pm 0.30}$
        & $3.26 \pm 2.00$ & $\mathbf{3.16 \pm 1.77}$
        & $10.28 \pm 2.45$ & $\mathbf{5.47 \pm 1.90}$
        & $17.8$ & $\mathbf{11.1}$
        \\

        XWALK$^\dagger$
        & -- & --
        & -- & --
        & $70.3$ & $\mathbf{161.5}$
        & $8.48 \pm 2.04$ & $\mathbf{1.88 \pm 0.14}$
        & $5.67 \pm 2.93$ & $\mathbf{4.24 \pm 1.67}$
        & $14.43 \pm 4.22$ & $\mathbf{6.30 \pm 1.71}$
        & $24.7$ & $\mathbf{10.0}$
        \\

        XBIKE$^\dagger$
        & -- & --
        & -- & --
        & $60.7$ & $\mathbf{144.5}$
        & $9.84 \pm 2.53$ & $\mathbf{1.95 \pm 0.17}$
        & $6.50 \pm 2.85$ & $\mathbf{4.75 \pm 1.77}$
        & $16.66 \pm 4.48$ & $\mathbf{6.92 \pm 1.86}$
        & $27.1$ & $\mathbf{11.6}$
        \\

        \bottomrule
    \end{tabular}%
    }
    \newline
    \scriptsize
    \textit{
    \textbf{ATE} $SE(3)$ aligned \quad
    \textbf{RTE} $\Delta = 6$ frames \quad
    \textbf{p99} 99th percentile of end-to-end frame time \quad
    \textbf{$\pm$}~frame time standard deviation \\
    * Numbers averaged over multiple runs \quad
    $\dagger$ long sequences that reach thermal limit (Sec.~\ref{sec:results}) 
    }
\end{table*}

We run all of our experiments on a \textbf{Samsung Galaxy S23} with a Snapdragon 8 Gen 2 SoC and a Hexagon V73 DSP. The device is powered by a 15\,Wh battery, and we disable cellular data and Wi-Fi to minimize background activity. The screen is dimmed to reduce power consumption further. We use the same camera calibration and configuration within each dataset family to ensure consistency across runs.

For the \textbf{dataset selection} we choose sequences spanning different sensors, environments, and motions. We evaluate on eight sequences from four distinct datasets: EuRoC \cite{burriEuRoCMicroAerial2016a} (an MAV), the Monado SLAM Dataset \cite{demayoMonadoSLAMDataset2025} (VR headsets), TUM-VI \cite{schubertBasaltTUMVI2018} (handheld rig), and a custom dataset recorded outdoors with a pair of XREAL AIR 2 Ultra glasses (wearable). We prefix them with \texttt{E}, \texttt{M}, \texttt{T}, and \texttt{X} respectively. Frames from these sequences and their corresponding sensor specifications are shown in the bottom of Fig.~\ref{fig:teaser}. The accuracy of our baseline system, Basalt, is benchmarked in \cite{usenkoBasaltVisualInertialMapping2020,demayoMonadoSLAMDataset2025}. We implement the same algorithm and, as Table~\ref{tab:performance} shows, reproduce its accuracy, so we do not repeat that evaluation here.

We compare the \textbf{\texttt{CPU}} baseline, the unmodified Basalt frontend built with -O3 for the Cortex-A715 and using OpenCV's and Eigen's NEON paths, against three DSP configurations, which we denote as \texttt{DSP}, \texttt{DSP$_B$}, and \texttt{DSP$_L$}. For all of them, we pin CPU threads to the mid SoC cluster, which we found to be the most energy-efficient for our workload. \textbf{\texttt{DSP}} is configured for maximum performance, with cDSP running at its \textsc{turbo} DCVS (Dynamic Clock and Voltage Scaling) corner at 1401.6\,MHz, using a polling FastRPC quality-of-service (QoS) policy, and a $40\,\mu$s DSP sleep-latency vote to keep the cDSP out of deep idle between calls. Under polling QoS the CPU spins until the call returns; under the adaptive policy it sleeps and is woken at the driver's predicted completion time. \textbf{\texttt{DSP$_B$}} is configured for maximum efficiency while maintaining the baseline's accuracy, with the cDSP running at the \textsc{SVS2} (Static Voltage Scaling) corner at 576\,MHz, using an adaptive FastRPC QoS policy, and the default sleep latency. \textbf{\texttt{DSP$_L$}} is similar to \texttt{DSP$_B$}, but we allow a decrease in accuracy by raising the FAST detection threshold floor from 5 to 20 and using 6 keyframes instead of 7 for the optimization.

\begin{figure}[t]
    \centering
    \includegraphics[width=\columnwidth]{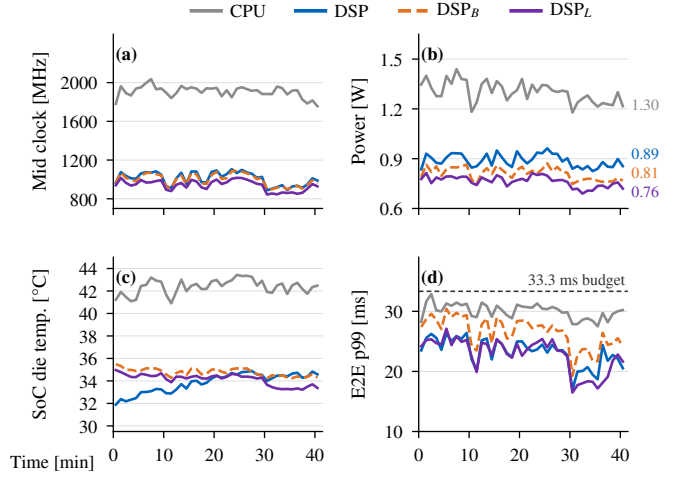}
    \caption{40-minute MOO12 run under CPU, DSP, DSP$_B$ and DSP$_L$: (a) pinned mid-cluster clock, (b) whole-device power from the sampled current and voltage (right-edge labels: run means; the charge counter quoted in the text reads 8--12\,\% higher), (c) SoC die temperature, the hottest on-die sensor, and (d) E2E p99 latency against the 33.3\,ms frame budget.}
    \label{fig:allday}
\end{figure}

We use \textbf{Perfetto} to record system traces for every run and get the timing, power, clock, and temperature numbers. Table~\ref{tab:power} uses the board thermistor adjacent to the SoC (\texttt{xo-therm}), while Fig.~\ref{fig:allday} reports the SoC die temperature, the hottest on-die \texttt{TSENS} zone. We use only \textbf{passive cooling}. At full throughput, both XREAL sequences in Table~\ref{tab:performance} thermally throttle the device, except XWALK on the \texttt{DSP}. We read the datasets from disk in chunks to avoid accessing the disk every frame; using the device camera \textbf{ISP} (image signal processor) via DMA in a real deployment would be more efficient. We run each configuration in two regimes: \emph{paced}, where replay follows the recorded timestamps so the pipeline must keep up with the sensor rate, and \emph{unpaced}, where the next frame is fed as soon as the pipeline accepts it, so each configuration reaches its maximum frame rate. In paced runs we count a frame as late when its end-to-end latency exceeds the frame period implied by the frame rate in Fig.~\ref{fig:teaser}.

We measure whole-device power draw from the on-device fuel gauge, with the device unplugged so that the counters record battery discharge. We report sampled current at Perfetto's default 250\,ms interval for Table~\ref{tab:power}, and integrated charge for Fig.~\ref{fig:allday}, which reads higher. An external power monitor is impractical on modern phones, so recent mobile energy studies measure from the internal gauge as we do~\cite{tuUnveilingEnergyEfficiency2023}. These gauges resolve power only at 100\,ms-1\,s, so we report no per-stage energy, and any systematic bias carries into the absolute figures. On the 40-minute MOO12 sequence, the two channels differ by 8--12\,\% in absolute power but agree within 2.4\,\% on the improvements. We state our results as improvements on the same device.

\begin{figure}[b]
    \centering
    \includegraphics[width=\columnwidth]{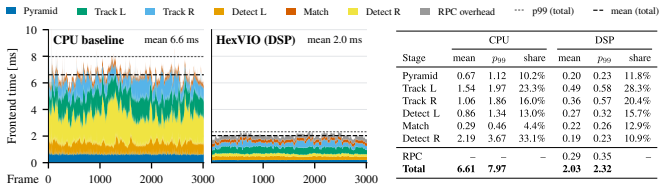}
    \caption{Per-stage frontend timing on MOO05, unpaced, both arms pinned to the mid cluster with four threads. Left: per-frame stage times of the CPU baseline and of HexVIO on the DSP at its maximum-performance operating point (TURBO, polling RPC), one fused dispatch per frame. Right: mean and 99th percentile per frame in ms and share of the summed stage means. RPC overhead, the round trip of the fused dispatch, exists only on the DSP side and counts in the DSP total but not in the shares.}
    \label{fig:timing}
\end{figure}

\textit{Performance.} Table~\ref{tab:performance} reports unpaced runs for maximum throughput. HexVIO is $1.65$--$2.38\times$ faster, most on the two thermally limited XREAL sequences. In Fig.~\ref{fig:timing}, we break down the frontend times for the MOO05 sequence from MSD and show that HexVIO improves times by $\sim 70\%$.

\begin{figure}[b]
    \centering
    \includegraphics{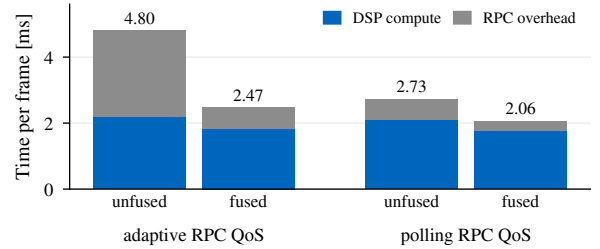}
    \caption{Effect of fusing the DSP calls on MOO05 (unpaced, TURBO, host on the mid cluster). Unfused issues six calls per frame (one stereo-pyramid build, two temporal tracks, two detections, one stereo track), fused one, under the adaptive QoS of DSP$_B$/DSP$_L$ and the polling QoS of DSP. Bars are DSP stage time plus the RPC round trip, as in Fig.~\ref{fig:timing}. Fusion saves 2.3\,ms per frame under adaptive QoS and 0.7\,ms under polling.}
    \label{fig:fusion}
\end{figure}

\textit{Fused dispatch.} Fig.~\ref{fig:fusion} isolates the benefits of the fused RPC dispatch: under the adaptive RPC policy, six per-stage calls cost 2.3\,ms per frame more than one fused call, almost all of it round-trip overhead, while the on-DSP time falls only slightly.

\paragraph*{Accuracy}
The DSP frontend preserves the baseline's accuracy. Across the six sequences with ground truth, ATE differs by at most 0.4\,\% between CPU and DSP on four of them. TR6 differs by 2.9\,\% in the DSP's favor (2.06 to 2.00\,cm) and EMH03 is 6\,\% worse (6.05 to 6.43\,cm). RTE is identical to two decimals on five of the six and differs by 0.01\,cm on MOO08. Frontend standard deviations shrink by a factor of five to fifteen (for example $9.84 \pm 2.53$ to $1.95 \pm 0.17$\,ms on XBIKE), which matters more than the mean for a system that must meet a per-frame deadline.

\begin{table}[t]
    \centering
    \caption{Power consumption on runs paced at sensor rate}
    \label{tab:power}
    \resizebox{\columnwidth}{!}{%
    \setlength{\tabcolsep}{2.5pt}
    \begin{tabular}{l *{16}{r}}

        \toprule

        & \multicolumn{4}{c}{\textbf{Power [W]}}
        & \multicolumn{4}{c}{\textbf{E2E [ms]}}
        & \multicolumn{4}{c}{\textbf{ATE [cm]}}
        & \multicolumn{4}{c}{\textbf{RTE [cm]}}
        \\

        \cmidrule(lr){2-5}
        \cmidrule(lr){6-9}
        \cmidrule(lr){10-13}
        \cmidrule(lr){14-17}

        \textbf{Sequence}
        & CPU & DSP & DSP$_{B}$ & DSP$_{L}$
        & CPU & DSP & DSP$_{B}$ & DSP$_{L}$
        & CPU & DSP & DSP$_{B}$ & DSP$_{L}$
        & CPU & DSP & DSP$_{B}$ & DSP$_{L}$
        \\

        \midrule

        EV102
        & 1.42 & 0.93 & 0.88 & \textbf{0.86}
        & 25.4 & 21.8 & 23.8 & \textbf{18.6}
        & 5.40 & 5.41 & 5.41 & \textbf{4.41}
        & \textbf{1.12} & \textbf{1.12} & \textbf{1.12} & \textbf{1.12}
        \\

        EMH03
        & 1.63 & 1.12 & 1.02 & \textbf{0.92}
        & 28.1 & \textbf{24.4} & 27.8 & 25.3
        & 6.05 & 6.43 & 6.43 & \textbf{6.01}
        & 1.18 & 1.18 & 1.18 & \textbf{1.17}
        \\

        TR2
        & 1.00 & 0.82 & 0.72 & \textbf{0.66}
        & 23.5 & 15.1 & 16.7 & \textbf{13.5}
        & \textbf{7.30} & 7.33 & 7.33 & 9.54
        & \textbf{0.70} & \textbf{0.70} & \textbf{0.70} & 0.80
        \\

        TR6
        & 1.07 & 0.86 & 0.77 & \textbf{0.66}
        & 23.6 & 17.8 & 18.9 & \textbf{14.2}
        & 2.06 & \textbf{2.00} & \textbf{2.00} & 3.30
        & \textbf{0.38} & \textbf{0.38} & \textbf{0.38} & 0.44
        \\

        MOO02
        & 1.23 & 0.84 & 0.77 & \textbf{0.72}
        & 18.0 & \textbf{13.0} & 15.0 & 13.1
        & 23.48 & \textbf{23.43} & \textbf{23.43} & 27.48
        & \textbf{1.54} & \textbf{1.54} & \textbf{1.54} & 1.59
        \\

        MOO08
        & 1.64 & 1.21 & 0.93 & \textbf{0.91}
        & 17.4 & \textbf{11.5} & 14.8 & 12.8
        & 2.74 & 2.73 & 2.73 & \textbf{2.67}
        & 1.12 & \textbf{1.11} & \textbf{1.11} & 1.12
        \\

        XWALK
        & 1.35 & 0.99 & 0.89 & \textbf{0.83}
        & 19.0 & 14.8 & 17.4 & \textbf{14.6}
        & -- & -- & -- & --
        & -- & -- & -- & --
        \\

        XBIKE
        & 1.31 & 0.99 & 0.90 & \textbf{0.84}
        & 18.9 & \textbf{14.8} & 17.3 & \textbf{14.8}
        & -- & -- & -- & --
        & -- & -- & -- & --
        \\

        \bottomrule
    \end{tabular}%
}

\vspace{0.25em}

\resizebox{\columnwidth}{!}{%
    \setlength{\tabcolsep}{2.5pt}
    \begin{tabular}{l *{16}{r}}

        & \multicolumn{4}{c}{\textbf{Energy [J]}}
        & \multicolumn{4}{c}{\textbf{E2E p99 [ms] / late [\#]}}
        & \multicolumn{4}{c}{\textbf{Temperature [°C]}}
        & \multicolumn{4}{c}{\textbf{Clock [MHz]}}
        \\

        \cmidrule(lr){2-5}
        \cmidrule(lr){6-9}
        \cmidrule(lr){10-13}
        \cmidrule(lr){14-17}

        \textbf{Sequence}
        & CPU & DSP & DSP$_{B}$ & DSP$_{L}$
        & CPU & DSP & DSP$_{B}$ & DSP$_{L}$
        & CPU & DSP & DSP$_{B}$ & DSP$_{L}$
        & CPU & DSP & DSP$_{B}$ & DSP$_{L}$
        \\

        \midrule

        EV102
        & 121 & 80 & 75 & \textbf{73}
        & 41.3/\textbf{0} & 35.4/\textbf{0} & 37.2/\textbf{0} & \textbf{32.4}/\textbf{0}
        & 32.0 & 30.0 & 29.5 & \textbf{29.4}
        & 1732 & 1188 & \textbf{1176} & 1196
        \\

        EMH03
        & 220 & 151 & 138 & \textbf{124}
        & 47.9/12 & \textbf{42.7}/\textbf{0} & 45.2/4 & 42.9/\textbf{0}
        & 33.7 & \textbf{29.6} & \textbf{29.6} & 29.7
        & 1932 & 1339 & 1250 & \textbf{1201}
        \\

        TR2
        & 144 & 118 & 104 & \textbf{96}
        & 33.4/\textbf{0} & 27.4/\textbf{0} & 27.7/\textbf{0} & \textbf{22.1}/\textbf{0}
        & 33.0 & 29.6 & 29.5 & \textbf{29.1}
        & 1318 & 1090 & 1148 & \textbf{956}
        \\

        TR6
        & 141 & 114 & 102 & \textbf{87}
        & 35.7/\textbf{0} & 30.0/\textbf{0} & 30.4/\textbf{0} & \textbf{24.0}/\textbf{0}
        & 32.0 & 29.9 & \textbf{29.5} & \textbf{29.5}
        & 1516 & 1112 & 1190 & \textbf{998}
        \\

        MOO02
        & 222 & 151 & 139 & \textbf{130}
        & 31.9/31 & 27.2/6 & 28.3/9 & \textbf{25.3}/\textbf{0}
        & 32.5 & 30.3 & 30.3 & \textbf{30.1}
        & 1887 & \textbf{989} & 1060 & 1013
        \\

        MOO08
        & 32 & 24 & \textbf{18} & \textbf{18}
        & 26.2/\textbf{0} & 23.1/\textbf{0} & 26.9/\textbf{0} & \textbf{22.5}/\textbf{0}
        & 34.2 & 31.2 & 28.9 & \textbf{28.5}
        & 1806 & 1257 & 1153 & \textbf{1134}
        \\

        XWALK
        & 988 & 723 & 655 & \textbf{607}
        & 32.5/128 & 25.7/\textbf{0} & 28.6/2 & \textbf{24.5}/1
        & 34.3 & 31.2 & 31.3 & \textbf{31.0}
        & 1853 & 1211 & 1176 & \textbf{1148}
        \\

        XBIKE
        & 1568 & 1183 & 1076 & \textbf{999}
        & 32.1/171 & 25.2/\textbf{0} & 28.0/6 & \textbf{24.4}/\textbf{0}
        & 33.5 & 31.7 & \textbf{31.4} & 31.7
        & 1863 & 1250 & 1218 & \textbf{1176}
        \\

        \bottomrule
    \end{tabular}%
}
    \newline
    \scriptsize
    \textit{
    \textbf{ATE} $SE(3)$ aligned \quad
    \textbf{RTE} $\Delta = 6$ frames \quad
    \textbf{E2E p99 / late} 99th percentile of frame times and late frame count \quad
    * Numbers are averaged over multiple runs
    }
\end{table}
\paragraph*{Efficiency}
Table~\ref{tab:power} pins every configuration to the sequence's sensor rate, so that all of them do the same work and the surplus throughput of Table~\ref{tab:performance} is spent on power instead of frame rate. DSP$_B$ is the same computation as DSP at a lower clock, so its accuracy is identical to DSP's; DSP$_L$ reaches a lower operating point by doing less work rather than by lowering the clock further.

At matched accuracy, DSP$_B$ draws 35\,\% less whole-device power than the CPU baseline on average, ranging from $-28$\,\% (TR2, TR6) to $-43$\,\% (MOO08); because the runs have equal duration, the energy reduction is the same. DSP$_L$ reaches $-40$\,\% on average and $-45$\,\% at best. Board temperature falls by 2--5\,°C on every sequence, and the pinned mid-cluster clock drops by up to 44\,\%, which is consistent with the power reduction: the CPU baseline must run the mid cluster near its top clock to meet the deadline, while the DSP meets it from a far lower operating point.

\paragraph*{Efficiency scaling}
Fig.~\ref{fig:teaser} shows the trade-off between power and throughput for both implementations on MOO05. Both curves start at the paced 30\,fps configuration and then run unpaced, so that each configuration reaches its maximum frame rate. The DSP curve dominates the CPU curve over its whole span: at the baseline's best throughput of 98\,fps (frontend pinned to the mid cluster, 4.8\,W) HexVIO needs 1.6\,W, and with only two host threads it already delivers 93\,fps at 1.5\,W. On MOO05, at the paced sensor rate, where both implementations meet the 30\,fps deadline, HexVIO draws 39\,\% less whole-device power. At the other end, the maximum-performance RPC policy lifts HexVIO to 180\,fps at 4.5\,W, a power level at which the CPU baseline delivers 97\,fps whether pinned or not: left to the scheduler it spends 7.0\,W for the same 97\,fps, because the prime and little cores join in. Pinning the CPU side of the pipeline to the mid cluster is therefore part of the design, not a benchmark artifact. On the DSP side the polling policy buys its throughput at a steep power price (Table~\ref{tab:power}), which is why DSP$_B$ and DSP$_L$ keep the driver defaults.

\paragraph*{Real-time behavior}
The clearest real-time result is in the late-frame columns of Table~\ref{tab:power} (bottom). Summed over the eight sequences the CPU baseline misses 342 deadlines, DSP$_B$ misses 21, and DSP$_L$ misses one. For the XREAL sequences, the larger numbers of late frames for the CPU baseline are not a throughput deficit: on frames that arrive on schedule, the CPU baseline misses only 0.16\,\%. The majority of late frames follow gaps in delivery, after which the governor lowers the mid-cluster clock. On XWALK, 81\,\% of late frames arrive after a gap, at a mean clock of 0.8\,GHz against 1.9\,GHz on schedule. The DSP frontend runs at a corner set from user space, so only the backend's end-to-end time is impacted by the host clock.

\paragraph*{All-day tracking}
We also show a long-term run of the 40-minute MOO12 sequence in Fig.~\ref{fig:allday}. Over the whole run the DSP configurations hold a lower and flat power draw, a stable mid-cluster clock and an SoC die temperature of 34--35\,°C (the CPU run holds 42\,°C), with at most 0.05\,\% late frames against 0.25\,\% for the CPU. The heat does not move to the accelerator: on the CPU run the hottest die sensor is in the mid cluster at 42\,°C and the idle cDSP reads 38\,°C, while under the DSP configurations the cDSP running the frontend stays at 33\,°C and no on-die sensor exceeds 35\,°C. Measured on the battery charge counter over the 72\,810 frames, the four configurations draw 1.46, 0.96, 0.90 and 0.83\,W (CPU, DSP, DSP$_B$, DSP$_L$) at the same accuracy (0.67--0.68\,m ATE, 7.7\,mm RTE). Given that this was run on a consumer smartphone with a 15\,Wh battery, 0.83\,W corresponds to about 18 hours of tracking, which we see as a promising step towards all-day tracking on commodity devices.

\section{CONCLUSIONS}

In this work, we ported the frontend of an open-source stereo VIO system to run on the DSP of an unmodified consumer phone, without root access and without changing the backend estimator. On a 40-minute sequence, we demonstrate that our tracking solution meets its frame deadline at 0.83\,W, which corresponds to about 18 hours on a 15\,Wh battery, while holding a flat mid-cluster clock and a 34--35\,°C SoC die temperature throughout, against 42\,°C for the CPU. Accuracy is preserved against the CPU reference; we reimplement the same algorithm. Fusing the per-stage dispatches into a single call per frame is what makes the efficiency operating points practical: with one call per frame instead of one per stage, the frontend can keep the power-saving RPC policy and still meet the deadline.

Future work can start with live sensor data. Qualcomm SoCs can stream the camera output directly into DSP L2~\cite{codrescuHexagon680DSP2015}, which would remove the DDR round trip for level~0. The interface sits inside the ISP pipeline and is available only to signed vendor code, so this is an OEM integration rather than a drop-in change. The backend optimizer is a second target, since it now accounts for most of the remaining frame time. Finally, Hexagon is not confined to smartphones: it ships in robotics boards such as the RB5 kit, in VR and AR headsets (XR2, AR1), and in automotive platforms (Snapdragon Ride), so a frontend that fits inside a phone's thermal budget carries over to the devices that spatial computing targets.

\bibliographystyle{IEEEtran}
\balance %
\bibliography{IEEEabrv,main}

\immediate\closein\imgstream

\end{document}